\documentclass[conference]{IEEEtran}
\IEEEoverridecommandlockouts
\usepackage{cite}

\usepackage{amsmath,amssymb,amsfonts}
\usepackage{url}
\usepackage{algorithmic}
\usepackage{graphicx}
\usepackage{textcomp}
\usepackage{xcolor,colortbl}
\usepackage{mathtools}
\usepackage{todonotes}
\usepackage{enumitem}
\usepackage[export]{adjustbox}
\usepackage{pifont}
\def\BibTeX{{\rm B\kern-.05em{\sc i\kern-.025em b}\kern-.08em
    T\kern-.1667em\lower.7ex\hbox{E}\kern-.125emX}}
\definecolor{Gray}{gray}{0.8}
\definecolor{LigthGray}{gray}{0.95}

\definecolor{CM10}{HTML}{008800}
\definecolor{CM09}{HTML}{00AA00}
\definecolor{CM08}{HTML}{00CC00}
\definecolor{CM07}{HTML}{00FF00}
\definecolor{CM02}{HTML}{A0FFA0}
\definecolor{CM01}{HTML}{C0FFC0}
\definecolor{CM00}{HTML}{EEFFEE}

\newcommand\mynobreakpar{\par\nobreak\@afterheading}

\usepackage[linesnumbered,ruled,vlined]{algorithm2e}
\usepackage{multirow}
\usepackage{array}
\usepackage{float}
\usepackage{acronym}
\usepackage{amsmath}

\newcolumntype{F}[1]{>{\columncolor{white}\raggedright}p{#1}}
\newcolumntype{S}[1]{>{\raggedright\arraybackslash\columncolor{LigthGray}}p{#1}}
\newcolumntype{L}{>{\columncolor{Gray}}l}
\newcolumntype{Q}{>{\columncolor{white}}l}
\newcolumntype{q}{>{\columncolor{white}}c}
\newcolumntype{a}{>{\columncolor{Gray}}c}
\newcolumntype{P}[1]{>{\centering\arraybackslash\columncolor{white}}p{#1}}
\newcolumntype{M}[1]{>{\centering\arraybackslash\columncolor{white}}p{#1}}
\newcolumntype{Z}[1]{>{\centering\arraybackslash\columncolor{LigthGray}}p{#1}}
\newcolumntype{X}{>{\columncolor{LigthGray}}c}
\newcolumntype{Y}{>{\columncolor{LigthGray}}l}
\newcolumntype{R}{>{\columncolor{LigthGray}}r}
\newcolumntype{O}[1]{>{\raggedleft\arraybackslash\columncolor{LigthGray}}p{#1}}

\makeatletter
 \let\old@ps@headings\ps@headings
 \let\old@ps@IEEEtitlepagestyle\ps@IEEEtitlepagestyle
 \def\confheader#1{%
 \def\ps@headings{%
 \old@ps@headings%
 \def\@oddhead{\strut\hfill#1\hfill\strut}%
 \def\@evenhead{\strut\hfill#1\hfill\strut}%
 }%
 \def\ps@IEEEtitlepagestyle{%
 \old@ps@IEEEtitlepagestyle%
 \def\@oddhead{\strut\hfill#1\hfill\strut}%
 \def\@evenhead{\strut\hfill#1\hfill\strut}%
 }%
 \ps@headings%
 }
\makeatother

\confheader{%
Submitted to The 19th International Conference on Network and Service Management (CNSM) 2023}

\begin{document}

\title{Network Traffic Classification based on Single Flow Time Series Analysis\\
\thanks{This research was funded by the Ministry of Interior of the Czech Republic, grant No. VJ02010024: Flow-Based Encrypted Traffic Analysis and also by the Grant Agency of the CTU in Prague, grant No. SGS23/207/OHK3/3T/18 funded by the MEYS of the Czech Republic.}
}

\author{\IEEEauthorblockN{1\textsuperscript{st} Josef Koumar}
\IEEEauthorblockA{
\textit{Czech Technical University in Prague}\\
Prague, Czech republic \\
koumajos@fit.cvut.cz
}
\and
\IEEEauthorblockN{2\textsuperscript{nd} Karel Hynek}
\IEEEauthorblockA{
\textit{CESNET, a.l.e.}\\
Prague, Czech republic \\
hynekkar@cesnet.cz, }
\and
\IEEEauthorblockN{3\textsuperscript{rd} Tomáš Čejka}
\IEEEauthorblockA{
\textit{CESNET, a.l.e.}\\
Prague, Czech republic \\
cejkat@cesnet.cz}
}

\maketitle

\begin{abstract}
    Network traffic monitoring using IP flows is used to handle the current challenge of analyzing encrypted network communication. Nevertheless, the packet aggregation into flow records naturally causes information loss; therefore, this paper proposes a novel flow extension for traffic features based on the time series analysis of the Single Flow Time series, i.e., a time series created by the number of bytes in each packet and its timestamp. We propose 69 universal features based on the statistical analysis of data points, time domain analysis, packet distribution within the flow timespan, time series behavior, and frequency domain analysis.  We have demonstrated the usability and universality of the proposed feature vector for various network traffic classification tasks using 15 well-known publicly available datasets. Our evaluation shows that the novel feature vector achieves classification performance similar or better than related works on both binary and multiclass classification tasks. In more than half of the evaluated tasks, the classification performance increased by up to 5~\%.  
\end{abstract}


\section{Introduction}
    Network traffic monitoring provides information about activities in a computer network---an essential insight for maintaining the service and its security. As the technology evolves, a classical approach using \textit{Deep Packet Inspection} (\textit{DPI}) is no longer feasible due to the increased privacy protection using encryption. Additional security features, such as the RFC draft  \textit{Encrypted Server Name Indication (ESNI)}~\cite{ietf-tls-esni-16}, which encrypts even domain names, forces the development of new ways of monitoring and analysis to detect network threats and malicious activities.  
    
    Contrary to DPI, flow-based~\cite{IP_flows} monitoring uses only aggregated information and statistics about the communication---IP flows. The IP flow term is defined, e.g.,  by \textit{Internet Protocol Flow Information Export (IPFIX)} specification as aggregated information about the sequence of packets observed within a specific timeslot with the same properties---usually IP addresses, transport protocol (often TCP or UDP), and ports. The most commonly used simple statistics are the sum of packets and the sum of bytes of the observed communication. Such representation of the traffic is universal enough to get a high-level overview of large networks with high volumes of traffic and even encrypted traffic.  
    
    Since flows contain mainly information from packet headers and do not extract the payload, they are not affected by the payload encryption and are the ideal candidate for encrypted traffic monitoring. Many research works~\cite{9892293,stergiopoulos2018automatic,8844609,wang2017malware,8844609,plny2023decrypto,montazerishatoori2020detection,doh_kamil,9375998} thus use it together with machine learning for encrypted traffic classification to increase visibility and identify encrypted malicious communication.

    Nevertheless, simple statistics such as the sum of transferred bytes and packets do not usually carry enough information for reliable traffic classification. The information about individual packet sizes, which has been found extremely useful in previous research works~\cite{luxemburk2022fine,sblt_our_paper,lashkari2017characterization}, is lost in the packet aggregation into flow records. Therefore, several approaches to extend flows  were proposed to increase the classification performance. For example, flows are often extended by the \textit{Sequence of packet lengths and times (SPLT)}~\cite{luxemburk2022fine} or \textit{Sequence of packet Burst Length and Time (SBLT)}~\cite{sblt_our_paper}, and application-specific information fields~\cite{http_extend_ip_flow,6356064,bayat2021deep}.

    The SPLT and SBLT sequences significantly increase the amount of information we can leverage for classification. Still, they cannot carry information about all packets transmitted in the flows for practical reasons such as limited memory of flow exporter or constrain on flow record size. Therefore, SPLT often contains only the first $n$ packets from the flow. For example, the Cisco joy exporter\footnote{\url{https://github.com/cisco/joy}} exports detailed information (packet size, timestamp, direction) up to the first 200 packets in a flow, ipfixprobe flow exporter\footnote{\url{https://github.com/CESNET/ipfixprobe}} exports this information for only the first 30 packets. 
    
    Even these packet-extended flows thus still miss a lot of information when dealing with longer communications. Therefore, our approach proposes an additional feature set to extend IP flows with \textit{Time Series Analysis (TSA)} to mitigate the information loss due to aggregation or limited SPLT or SBLT sequence size. Instead of extending flows for information about individual packets, we extend flows for  69 novel features and test them for network traffic classification. In our approach, we consider the flow as time series of network packets, i.e., Single Flow Time Series (SFTS). Using the analysis of SFTS, we generate a set of significant features, which describes time dependencies between packets, packet sequences, distribution of packets, and behavior of packets. We evaluate the usability and universality of the feature set using 23 different network classification tasks with 15 well-known public datasets and machine learning algorithms. Our evaluation showed that the novel feature vector achieves excellent classification performance, similar to or better than related works in both binary and multiclass classification tasks. In more than half of the evaluated tasks, the classification performance increased by up to 5~\%. 

    Furthermore, we also performed feature reduction to enable the deployment on networks with size-constrained network telemetry (e.g., due to available bandwidth allocated for monitoring). Despite the decrease in available information, the reduced feature vector of only ten features still achieves very good performance and reduces the average classification accuracy (compared to the full feature vector of 69 features) by only 0.03\%.  

    The main contributions of our work can be summarized as follows:

    \begin{itemize}
        \item[--] We proposed a novel approach that uses Time Series Analysis to generate 69 novel features. 
        \item[--] We computed the proposed feature vector for 15 well-known network datasets and made them publicly available at Zenodo platform~\cite{josef_koumar_2023_8035724}. 
        \item[--] Using the novel features, we designed network classifiers capable of multiple potential network threat detection using machine learning algorithms. The novel classifiers achieved excellent accuracy, exceeding the previous best results from relevant works. Threats include Botnet, Cryptomining, DoH, (D)DoS, Malicious DNS, Intrusion in IDS, IoT Malware, Tor, and VPN.
        \item[--] Using the novel features, we designed several multiclass classifiers, which performed better than previously published state-of-the-art algorithms. The multiclass classification concerns Botnet, IDS, IoT Malware, Tor, and VPN. 
        \item[--] We performed feature reduction to 10 significant features. The reduced feature vector still achieves very high accuracy across the evaluated classification tasks. The average loss across the evaluated tasks compared to the full feature vector is equal to  $-0.09\%$ (std $\pm$~$0.25\%$)  F1-score for binary classification, and $-0.19\%$  (std $\pm$~$0.99\%$)  weighed F1-score for multiclass classification.
    \end{itemize}

    This paper is divided as follows: Section~\ref{related_works_section} summarizes the related work of flow-based network traffic classification. Section~\ref{tsa_in_flow_exporter} provides information about time series analysis concepts and describes a novel approach to time series analysis in the IP flow exporter. Section~\ref{exported_features} provides a complete description of features exported in the novel extended IP flow. Section \ref{classification_section} describes the complete classification pipeline with classification results. Section~\ref{feature_reduction_section} contains the results of feature reduction. Section~\ref{conclusion_section} concludes this paper.

\section{Related works} \label{related_works_section}
    Flow-based network classification is an important area with multiple challenging tasks and various approaches. The main constraint of the detection method lies in the input data and information extracted by the monitoring system. For example, the flow monitoring systems based on NetFlowV5\footnote{\url{https://www.cisco.com/c/en/us/td/docs/net_mgmt/netflow_collection_engine/3-6/user/guide/format.html}} can export only basic statistics about the ongoing communication, significantly constraining the subsequent network detectors that often need additional data sources to maintain reasonable accuracy~\cite{Jerabek2022}. Thus, many proposals extend the basic flow records for various information. We can divide the flow extension into two main approaches: 1) Extension for packet sequences and 2) Extension for precomputed features.

\subsection{Extension for raw packet information}
    The extension of flows for packet sequences embeds the raw packet-level information about ongoing connections into the flows. Typically, flows are extended for a sequence of packets lengths and times (SPLT) that can be directly used for classification as in the case of Luxemburk et al.~\cite{luxemburk2022fine}, or can be additionally processed for additional feature extraction as in the case of~\cite{vekshin2020doh}. 

    Nevertheless, the SPLT sequence cannot contain data about all packets in the flows due to practical reasons. The larger flows require more processing power and consume more memory and bandwidth. Thus the ipfixprobe flow exporter limits the size of the SPLT sequence to 30 packets.

    To capture information about the packets that do not fit into the SPLT, researchers extend flows for additional features that we consider raw. For example, Tropkova et al.~\cite{sblt_our_paper} proposed to use a Sequence of Burst lengths and Times (SBLT), which carries the information about individual packet bursts (times, amount of transferred data). Nevertheless, even SBLT has its length limit. Moreover, the aggregation of packets into the bursts loses some information about the exact timing of packets inside the burst. 

    Hofstede et al. used traffic histograms~\cite{Hofstede2017} containing the distribution of packet lengths across the whole flow. Similarly, as SBLT, histograms do not carry information about the timing of individual packets. 

\subsection{Extension for precomputed features}
    Instead of exporting raw packet data that can be then processed by additional feature extraction, this approach computes the statistical features inside the exporter itself. An example of such an exporter is the CICFlowMeter\footnote{\url{https://github.com/ahlashkari/CICFlowMeter}} that extends each flow with 80 statistical features---mainly mean, standard deviation, max, and min of multiple countable information from packets, such as the number of packets and bytes. These features are then used by multiple researchers in various network classification tasks~\cite{9796558,cic_bell_dns_2021_article,sharafaldin2018toward,Agrafiotis_Makri_Flionis_Lalas_Votis_Tzovaras_2022,ding2022imbalanced}.
    
    Similarly, as CICFlowMeter, MontazeriShatoori et al.~\cite{DoH_cic_dataset} created a DoHLyzer exporter\footnote{\url{https://github.com/ahlashkari/DoHLyzer}} that produces features directly within the flows. Nevertheless, the feature vector is entirely different from the features supported by the CICFlowMeter. 

    Compared to the SPLT and other raw-packet flow extensions, the computation of features directly in the exporter can capture statistical information across the whole flow, and no packet is missed. Nevertheless, it also aggregates packets. The packet aggregation then causes information loss, especially in the timing domain, which is not properly captured by the existing exporters and their feature extraction capability. However, time-related features such as periodicity are essential in the network classification, as shown by Koumar et al.~\cite{koumar2022network} or MontazeriShatoori et al.~\cite{DoH_cic_dataset}.

    In this work, we focus on the IP-flow extension for precomputed features to capture information about all packets. Compared to previous approaches, we aimed to create a universal feature vector that contains features based on statistical, time, distribution, frequency, and behavior properties acquired from the Time Series Analysis of the packet time series of each flow, i.e., SFTS~\cite{KoumarUSTS}. Moreover, compared to all previous approaches, the universality and usefulness of the feature vector have been verified on 23 different network classification tasks using 15 network datasets. 

\section{Time Series Analysis in the IP Flow Exporter} \label{tsa_in_flow_exporter}
    This section describes several terms to explain our approach to time series analysis in the flow exporter to create a novel flow extension. We consider two crucial times for exporting flows---the inactive timeout is set to 65 seconds, and the active timeout is set to 300 seconds\footnote{If no packet is observed within the ``inactive timeout'' period, the flow is considered terminated. Flows longer than the ``active timeout'' are split and are exported every time this timeout elapses.}. These settings belong to the open source IP flow exporter \textit{ipfixprobe}.

    In the state of the art of analysis of the time series from network traffic are mostly time series considered with evenly spaced time between observations. This type of time series is called evenly spaced or regularly sampled, and it is defined as the sequence of observation \(\{X_n\} = \{x_1, \dots, x_n\}\) taken in times \(\{T_n = t_1, \dots, t_n\}\), where \(n\) is the number of observations and is always true: \(t_{j+1} - t_{j} = t_{j} - t_{j-1}, \forall j \in {2, \dots, n}\). Because of this behavior, it is possible to apply subtraction and division and get the sequence of times \(\{T_n\} = {1,2, \dots, n}\). So when an evenly spaced time series is used, then it is written only as \(\{X_n\}\) where \(n := 1,2,\dots,n\) and absolute observation times are unnecessary.  

    It is possible to use evenly spaced time series to analyze network traffic, mainly for forecasting and anomaly detection. Some previous works \cite{montazerishatoori2020detection} use evenly spaced time series even for classification. However, network traffic naturally occurs with unevenly spaced timestamps (packet transmission time). Moreover, to create an evenly spaced time series, we need to set the aggregation interval---the time window for a single datapoint in the series---that highly affects the analysis result due to packets occurring at the aggregation interval borders. Badly selected aggregation intervals then cause analysis failure. Unfortunately, each time series has a different ideal aggregation interval---thus, the analysis failure with evenly spaced time series is (for some time series) inevitable~\cite{KoumarUSTS}. 
    
\begin{figure}[htbp]
            \centerline{\includegraphics[width=9cm]{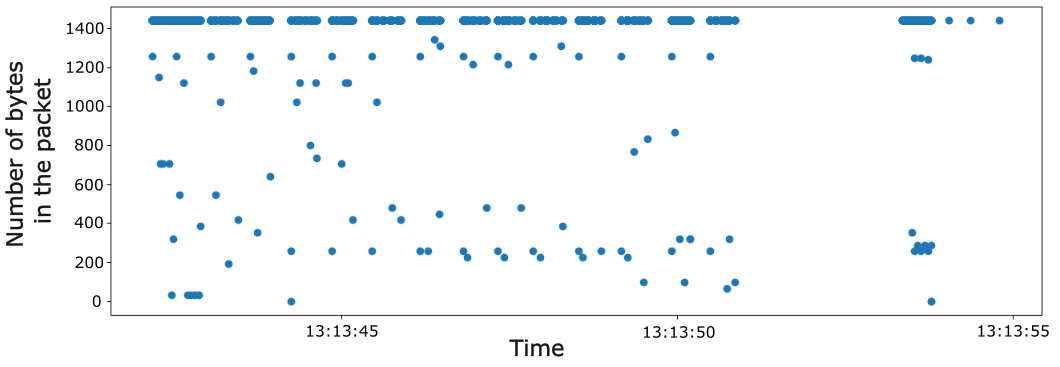}}
            \caption{Example of Single Flow Time Series (SFTS). It is possible to see that most of the packets are sent in close succession with the number of bytes equal to the Maximum Transmission Unit (MTU) of Ethernet, i.e., 1500 bytes. }
            \label{plot:packet_time_series}
    \end{figure}

    In our approach, we create time series from packets within a flow---the series payload sizes in bytes with the corresponding transmission timestamp to create a time series. We call them Single Flow Time Series (SFTS). However, the SFTS created by the sizes of packets and their timestamps do not have evenly spaced timestamps between the datapoints. That means a time series of observations \( \{X_n\} = \left(x_1, x_2, \dots, x_n\right)\) taken at times \( \{T_n\} = \left(t_1, t_2, \dots, t_n\right)\) does not have constant \(\delta_j = t_{j+1} - t_{j}, \forall j \in \{1,\dots, n-1\}\). This type of time series is called unevenly (or unequally/ irregularly) spaced. The example of SFTS is shown in Figure \ref{plot:packet_time_series}. The SFTS and the non-evenly-spaced time analysis methods are used in this study to create a novel feature vector.

\section{Features description} \label{exported_features}
    This section contains a detailed description of novel time series features. We organized the features into five categories: 1)~statistical, 2)~time-based, 3)~frequency-based, 4)~distribution based, and 5)~behavioral. Some of our proposed features for network classification were already used for classification in other fields of science, such as music classification \cite{SZABO200497,scheirer1997construction,lerch2012introduction}.  The detailed description with mathematical equations of the whole feature set is published on the Zenodo platform~\cite{josef_koumar_2023_8035724}.
    
    \subsection{Statistical-based features}\label{seq:statistical_based}
        The first set of features is based on statistical evaluation of the sequence of observation \(\{X_n\}\) of the SFTS. The idea is a statistical description of data point deviation, i.e., statistical deviation of the packets' payload lengths. Table \ref{tab:statistical_features} shows the list of statistical-based features.

        \begin{table*}
            \caption{List of statistical-based features}
            \begin{center}
                \begin{tabular}{|Y|Y|Y|Y|Y|Y|Y|}
                    \hline
                    \rowcolor{Gray}\textbf{Feature} & \textbf{Feature} & \textbf{Feature} & \textbf{Feature}  & \textbf{Feature}  & \textbf{Feature} & \textbf{Feature} \\
                    \hline
                    \rowcolor{white} \textit{Mean} & \textit{Median} & \textit{Standard deviation} & \textit{Percent above mean} & \textit{Fisher-Pearson \(G_1\) skewness} & \textit{Coefficient of variation} & \textit{Kurtosis} \\
                    \hline
                    \textit{Variance} & \textit{Burtiness} & \textit{First quartile} & \textit{Percent below mean} & \textit{Fisher-Pearson \(g_1\) skewness} & \textit{Pearson SK\(_1\) skewness} & \textit{Entropy} \\
                    \hline
                    \rowcolor{white} \textit{Third quartile} & \textit{Min} & \textit{Max} & \textit{Min minus max} &  \textit{Fisher \(\mu_3\) skewness} & \textit{Scaled entropy} & \\
                    \hline
                    \textit{Percent deviation} & \textit{Mode} & \textit{Average dispersion} & \textit{Root mean square} & \textit{Pearson SK\(_2\) skewness} & \textit{Galton skewness} & \\
                    \hline
                \end{tabular}
                \label{tab:statistical_features}
            \end{center}
        \end{table*}

    \subsection{Time-based features}\label{seq:time_based}
        The time-based features describe the time axis of the unevenly-spaced time series \(\{x_n\}\). For computation time-based features, we use a sequence of relative times \(\{rt_{n}\} = t_{i} - t_{0}, i \in \{1,\dots,n\}\), i.e., time from the beginning of a flow. Additionally, we use the sequence of time differences \(\{dt_{n-1} = t_{i+1} - t_{i}, i \in \{1,\dots,n-1\} \}\), i.e., time spaces between packets. The set of time-based features that are exported in the extended flow is listed below:
        
        \begin{description}
            \item[Mean, median, 1$^{st}$, and 3$^{rd}$ quartile of relative times] features are computed from the relative times $rt_{n}$ to capture the statistical properties of packet times.
            \item[Mean, median, min, and max of time differences] features are statistics of the time differences $dt_{n}$ and represent information about spaces in the SFTS of the flow. 
            \item[Duration] is the last data point in the relative times $rt_n$.
        \end{description}
        
    \subsection{Distribution-based features}\label{seq:distribution_based}
        The set of distribution-based features that are exported in the extended flow describes the distribution of data points in the SFTS \(\{x_n\}\). The distribution features are listed below:

        \begin{description}
            \item[Hurst exponent] can identify three behavior of time series---long-term switching between high and low values, long-term autocorrelation, and random (uncorrelated) time series~\cite{hurst_exponent}.
            \item[Stationarity] indicates the stationarity of the time series.
            \item[Benford's law] computes the probability of satisfaction of Benford's law~\cite{miller2015benford} for occurrence counts of the nine most frequent packet lengths.
            \item[Normal distribution] captures the probability that the SFTS is distributed by the normal distribution.
            \item[Count distribution] captures the packet distribution within the SFTS---if the majority of data was sent at the begging or at the end.
            \item[Count non-zero distribution] is similar to feature \textit{Count distribution} but filters the data points with zero value.
            \item[Time distribution] describes the deviation of time differences between individual packets within the SFTS.
        \end{description}

    \subsection{Frequency-based features}\label{seq:frequency_based}
        The idea of frequency-based features is to transform time series into the frequency domain and analyze it. Based on recent research~\cite{fda_tda_comparsion_1,fda_tda_comparsion_2,fda_tda_comparsion_3}, the frequency domain has several advantages over the time-domais. Frequency domain analysis 1)~allows for a more compact representation of a time series, 2)~is particularly useful for analyzing periodic behaviors because it allows analysis of the individual frequency components, 3)~can be used to compare the frequency content of different time series, that can be useful for identifying similarities or differences between time series and common features or patterns in a set of time series, 4)~can filter out unwanted frequency components from a time series that is useful for eliminating noise or other unwanted artifacts from a time series, and lastly 5)~can help to identify the underlying sources of variation in a time series. So it is possible that we can get suitable features for the classification of network traffic from the frequency domain.
        
        Since the SFTS are unevenly spaced, we must use the Lomb-Scargle (LS) periodogram~\cite{lomb,scargle,LS_periodogram} to transform the time series into a frequency domain. LS was originally developed for unevenly spaced time series in astrophysics.  
        
        The set of frequency-based features is listed below:

        \begin{description}
            \item[Min power, Max power] features represent the minimum and maximum power of the LS periodogram.
            \item[Frequency of min power, Frequency of max power] features describe the frequency of the minimum and maximum power of the LS periodogram.
            \item[Power mode, mean, stdev] features describe the statistics of the power spectrum of the LS periodogram.
            \item[Spectral bandwidth] describes the difference between upper and lower frequencies~\cite{SZABO200497}. 
            \item[Spectral centroid] indicates at which frequency the energy of a spectrum is centered upon~\cite{scheirer1997construction}.
            \item[Spectral energy] represents the total energy present at all frequencies in LS periodogram.
            \item[Spectral entropy] is the degree of randomness or disorder in the LS periodogram.
            \item[Spectral flatness] estimates the uniformity of signal energy distribution in the frequency domain~\cite{boashash2015time}.
            \item[Spectral flux] is the rate of change of periodogram power with increasing frequency~\cite{scheirer1997construction}.
            \item[Spectral kurtosis] can indicate a non-stationary or non-Gaussian behavior in the power spectrum~\cite{spectral_kurtosis_compute}.
            \item[Spectral periodicity] decides if in the LS periodogram is a significant peak that indicates the periodicity~\cite{about_scdf}. 
            \item[Spectral rolloff] is defined as frequency bellow at is concentrated \(85\%\) of the distribution power ~\cite{spectral_rolloff_zero_cros}.
            \item[Spectral spread] is the difference between the highest and lowest frequency in the power spectrum~\cite{peeters2004large}.
            \item[Spectral skewness] is the measure of peakedness or flatness of power spectrum~\cite{peeters2004large}. 
            \item[Spectral slope] is the slope of the power spectrum trend in a given frequency range~\cite{lerch2012introduction}. 
            \item[Spectral zero crossing rate] refers to the rate of power shifts, i.e., the change from negative to positive or the reverse~\cite{spectral_rolloff_zero_cros}. 
        \end{description}

    \begin{table*}
        \caption{Summarized related works for classification. If the ``--'' appears, then the related works do not present the metrics, or the dataset is not designed for multiclass classification.}
        \begin{center}
            \begin{tabular}{|Q|Q|M{1.2cm}|M{1.2cm}|M{0.2cm}|Q|M{1.2cm}|M{1.2cm}|M{1.2cm}|}
                \cline{1-4} \cline{6-9}
                \cellcolor{Gray} & \multicolumn{3}{c|}{\cellcolor{Gray}\textbf{Binary classification}} & & \multicolumn{4}{c|}{\cellcolor{Gray}\textbf{Multiclass classification}}\\
                \cline{1-4} \cline{6-9}
                \cellcolor{Gray}\textbf{Detection problem} & \cellcolor{Gray} \cellcolor{Gray} \textbf{Method} & \cellcolor{Gray} \textbf{Accuracy} & \cellcolor{Gray} \textbf{F1-score} & & \cellcolor{Gray} \textbf{Method} & \cellcolor{Gray} \textbf{Average} & \cellcolor{Gray} \textbf{Accuracy} & \cellcolor{Gray} \textbf{F1-score} \\
                \cline{1-4} \cline{6-9}
                 &   Stergiopoulos  et  al.  ~\cite{stergiopoulos2018automatic}  & 99.85 &  99.90  & & Marín et al.~\cite{8844609}  & macro & 99.72 & 76.04 \\
                \cline{2-4} \cline{6-9}
                \multirow{-2}{*}{CTU-13~\cite{GARCIA2014100}} & Wang et al.~\cite{wang2017malware}   & 99.41 & $\thicksim 99$ & &  Gonzalo et al.~\cite{8844609} & macro & 76.47 & 76.17  \\ 
                \cline{1-4} \cline{6-9}
                CESNET-MINER22~\cite{richard_plny_2022_7189293} & Plný et al.~\cite{plny2023decrypto}  & 93.72 &  90.59  & & \multicolumn{4}{q|}{--} \\
                \cline{1-4} \cline{6-9}
                & Kumaar et al.~\cite{kumaar2021hybrid}  & 99.19 & 99.20 &  & \multicolumn{4}{q|}{--} \\
                \cline{2-4} \cline{6-9}
                \multirow{-2}{*}{CIC-Bell-DNS~\cite{cic_bell_dns_2021_article}} & Mahdavifar  et  al.  ~\cite{cic_bell_dns_2021_article}  &	98.9 & 98.9 &  & \multicolumn{4}{q|}{--} \\
                \cline{1-4} \cline{6-9}
                &   Zebin et al.~\cite{9796558}  & 99.98 & 99.91   & &\multicolumn{4}{q|}{--} \\
                \cline{2-4} \cline{6-9}
                \multirow{-2}{*}{CIC-DoHBrw-2020~\cite{montazerishatoori2020detection}} &  MontazeriShatoori~\cite{montazerishatoori2020detection}  & -- &  99.30   & & \multicolumn{4}{q|}{--} \\
                \cline{1-4} \cline{6-9}
                DoH-Real-world~\cite{Jerabek2022}  & Jeřábek et al.~\cite{doh_kamil} &  97.5 &  98.7 &  & \multicolumn{4}{q|}{--} \\ 
                \cline{1-4} \cline{6-9}
                HTTPS Brute-force~\cite{jan_luxemburk_2020_4275775} & Luxemburk et al.~\cite{9375998}   & 99.93 &  96.26 &  & \multicolumn{4}{q|}{--} \\
                \cline{1-4} \cline{6-9}
                &  Shafiq  et  al.~\cite{SHAFIQ2020433}  & 99.99  & 99.99  &  & \multicolumn{4}{q|}{--} \\
                \cline{2-4} \cline{6-9}
                \multirow{-2}{*}{Bot-IoT~\cite{dos_iot_Dataset}} & Shafiq  et  al.~\cite{SHAFIQ2020101863}  & 99.99 &  99.99  &  & \multicolumn{4}{q|}{--} \\ 
                \cline{1-4} \cline{6-9}
                &  Khacha et al.~\cite{khacha2022hybrid}   & 99.99  &   99.99 &  & Khacha et al.~\cite{khacha2022hybrid}  &    weighted & 98.69 & --   \\
                \cline{2-4} \cline{6-9}
                \multirow{-2}{*}{Edge-IIoTset~\cite{mbc1-1h68-22}} & Ferrag et al.~\cite{9751703}     & 99.99 &  99.96 &  & Tareq et al. ~\cite{tareq2022analysis}  & weighted & 94.94 & -- \\
                \cline{1-4} \cline{6-9}
                &  Sahu et al.~\cite{sahu2021internet} & $\thicksim 96$ &   $\thicksim 96$  &  & \multicolumn{4}{q|}{--} \\
                \cline{2-4} \cline{6-9}
                \multirow{-2}{*}{IoT-23~\cite{sebastian_garcia_2020_4743746}} &    Nascita et al.~\cite{nascita2022machine}   & 99.93 &  91.70   &  & \multicolumn{4}{q|}{--} \\
                \cline{1-4} \cline{6-9}
                &  Dai et al. ~\cite{dai2023glads}  & 99.29  &  99.03  & & Tareq et al. ~\cite{tareq2022analysis}  & weighted & 98.5 &  98.57  \\
                \cline{2-4} \cline{6-9}
                \multirow{-2}{*}{TON\_IoT~\cite{moustafa2021new} } & Guo~\cite{guo2021machine}  & 99.23 &  98.90  &  & Dai et al. ~\cite{dai2023glads}  & weighted & 98.18  & 95.12 \\
                \cline{1-4} \cline{6-9}
                &  Agrafiotis~\cite{Agrafiotis_Makri_Flionis_Lalas_Votis_Tzovaras_2022}  &  98.5 &   95.4 & &  Kunang et al.  & weighted & 95.79 & 95.11 \\
                \cline{2-4} \cline{6-9}
                \multirow{-2}{*}{CIC-IDS-2017~\cite{sharafaldin2018toward} } &    Ding et al.~\cite{ding2022imbalanced}  & 95.86 & 95.81 & & Leon et al. ~\cite{9892293}   & micro & 99.86 & --  \\
                \cline{1-4} \cline{6-9}
                &   Ding et al.~\cite{ding2022imbalanced}   & 92.39 & 94.39 & & Ding et al. ~\cite{ding2022imbalanced}  & macro & 90.39 & 79.64  \\
                \cline{2-4} \cline{6-9}
                \multirow{-2}{*}{UNSW-NB15~\cite{moustafa2015unsw} } &   Mulyanto et al.~\cite{mulyanto2020effectiveness}  & 86.73 & 90.41 &  & Mulyanto et al.~\cite{mulyanto2020effectiveness}    & macro & 73.39 &  39.78  \\
                \cline{1-4} \cline{6-9}
                & Sarkar et al.~\cite{sarkar2020detection}  & 99.89 & 99.88   & & Yang et al. ~\cite{yang2023network}  & weighted & 96.04 & 95.97 \\
                \cline{2-4} \cline{6-9}
                \multirow{-2}{*}{ISCX-Tor-2016~\cite{lashkari2017characterization}}  & Cuzzocrea et al.~\cite{cuzzocrea2017tor}  & -- &  99.5  & &  Dai et al. ~\cite{dai2023glads}   & weighted & 97.95 &  86.77  \\
                \cline{1-4} \cline{6-9}
                & Aceto  et  al.~\cite{ACETO2021102985}  & 93.75  & 91.95  & & Dener et al.  ~\cite{dener2023rfse} & macro & 89.29 &  87.83  \\
                \cline{2-4} \cline{6-9}
                \multirow{-2}{*}{ISCX-VPN-2016~\cite{icxs_vpn_2016_dataset}} &  Shapira et al.~\cite{shapira2019flowpic}   & 88.4 & -- &  & Aceto  et  al.~\cite{ACETO2021102985}  & macro & 73.14 & 71.14  \\
                
                \cline{1-4} \cline{6-9}
                & Jorgense et al. ~\cite{vnet_dataset}  & -- & 98.00  &  &  Jorgense et al. ~\cite{vnet_dataset}  & micro & 96 &  96  \\
                \cline{2-4} \cline{6-9}
                \multirow{-2}{*}{VNAT~\cite{vnet_dataset}} &  Holodnak et al.~\cite{10029386}  &  -- & 85.30  & &  Holodnak et al.~\cite{10029386}  & macro & -- & 85.3  \\
                \cline{1-4} \cline{6-9}
            \end{tabular}
            \label{tab:related_works}
        \end{center}
     \end{table*}

    \subsection{Behavior-based features}\label{seq:behavior_based}
        The behavior-based features are focused on describing the specific set of behaviors of the SFTS. The set of behavior-based features that are exported in the extended flow is listed below:

        \begin{description}
            \item[Significant spaces] indicates if there are some significantly bigger spaces between packets.
            \item[Switching ratio] represents a value change ratio (switching) between payload lengths.
            \item[Transients] indicates if at least one transient in the time series exists. The transient is the behavior when a set of data points occurring in a short time window has significantly larger values.
            \item[Count of zeros] represents a percentage of one-second intervals that do not contain any packets.
            \item[Biggest interval] contains the maximal amount of data transferred in a one-second interval.
            \item[Directions] describes a percentage ratio of packet direction. 
            \item[Periodicity] is the length and time of periodically occurring packet, if present.
        \end{description}

\section{Dataset Selection}
    We explored multiple publicly available datasets previously used or published in the network traffic classification domain. Nevertheless, a lot of datasets consist of already precomputed features and do not contain raw packet-based data, which is necessary for our feature extraction based on time-series analysis. Thus, we considered mainly the datasets where raw packet captures (PCAP files) were available. Together we selected 15 well-known network datasets that are written in Table~\ref{tab:related_works} and processed them with our feature extraction. The processed datasets with our feature set were also published at Zenodo~\cite{josef_koumar_2023_8035724}. 
    
    The selected datasets cover the most important traffic detection (binary) or classification (multiclass) tasks: 1)~Botnet detection/classification, 2)~Cryptomining detection, 3)~DNS malware detection, 4)~DNS over HTTPS detection, 5)~DoS attack detection, 6)~HTTPS Bruteforce detection, 7)~Intrusion detection/classification, 8)~IoT malware classification, 9)~TOR detection/classification and 10)~VPN traffic detection/classification.
    
    In order to evaluate the performance of the novel features, we needed to create the baseline---a best-performing classifier for each concerned dataset. We searched for recent classifier proposals (published after 2017) using public research paper databases such as Google Scholar, IEEE Explore, and ACM Digital Library. We went through more than 300 papers and selected the best-performing proposals that met the following conditions ensuring fair comparability: 1)~it was a flow-based method, 2)~it uses the dataset as a whole and classifies all the datasets classes and types of samples, 3)~does not use IP addresses as input features\footnote{The concerned datasets are mainly lab-created; thus usage of IP addresses is not considered---in this case---as a good practice due to dataset overfitting as described by Behnke et al.~\cite{behnke2021feature}}, 4)~does not combine the concerned dataset with additional data. The selected best-performing proposals for both binary or multiclass versions of the classification tasks for each dataset are written in Table~\ref{tab:related_works}.

   \begin{table*}
                \caption{Final results (in \%) of the testing phase of binary classification based on the TSA of SFTS. Results contain achieved scores of the best model from the validation phase on the test dataset, and they are compared with the best results from related works. The green-colored cells represent results where our approach is significantly (by 1\% or more) better than best-related work. Contrary the red-colored cells represent results where our approach is significantly worse than best-related work. Furthermore, the gray-colored cells represent results that are similar to best-related work.}
                \begin{center}
                    \begin{tabular}{|Q|Y|Z{1.1cm}|Z{1.1cm}|F{0.5mm}|Y|Y|Z{1.1cm}|Z{1.1cm}|}
                        \cline{1-4} \cline{6-9}
                        \cellcolor{Gray} & \multicolumn{3}{c|}{\cellcolor{Gray}\textbf{Binary classification}} & & \multicolumn{4}{c|}{\cellcolor{Gray}\textbf{Multiclass classification}}\\
                        \cline{1-4} \cline{6-9}
                         \cellcolor{Gray}\textbf{Detection problem} & \cellcolor{Gray} \cellcolor{Gray} \textbf{Method} & \cellcolor{Gray} \textbf{Accuracy} & \cellcolor{Gray} \textbf{F1-score} & & \cellcolor{Gray} \textbf{Method} & \cellcolor{Gray} \textbf{Average} & \cellcolor{Gray} \textbf{Accuracy} & \cellcolor{Gray} \textbf{F1-score} \\
                        \cline{1-4} \cline{6-9}
                         & Stergiopoulos  et  al.  ~\cite{stergiopoulos2018automatic}  & 99.85  & 99.90 & & \cellcolor[HTML]{BBFFBB}Marín et al.~\cite{8844609}  & \cellcolor[HTML]{BBFFBB}Macro & \cellcolor[HTML]{BBFFBB}99.72  & \cellcolor[HTML]{BBFFBB}76.04 \\
                        \cline{2-4} \cline{6-9}
                        & & & &  & \cellcolor[HTML]{BBFFBB} &  \cellcolor[HTML]{BBFFBB}Macro &  \cellcolor[HTML]{BBFFBB}\textbf{99.73} & \cellcolor[HTML]{BBFFBB}\textbf{82.79} \\
                        \multirow{-3}{*}{\textbf{Botnet}} & \multirow{-2}{*}{Our approach} & \multirow{-2}{*}{99.98} & \multirow{-2}{*}{99.93}  &  & \cellcolor[HTML]{BBFFBB}\multirow{-2}{*}{Our approach} & \cellcolor[HTML]{BBFFBB}Weighted & \cellcolor[HTML]{BBFFBB}\textbf{99.73} & \cellcolor[HTML]{BBFFBB}\textbf{99.73} \\
                        \cline{1-4} \cline{6-9}
                         & \cellcolor[HTML]{BBFFBB}Plný et  al.  ~\cite{plny2023decrypto} &  \cellcolor[HTML]{BBFFBB}93.72 & \cellcolor[HTML]{BBFFBB}90.59  &  & \multicolumn{4}{c|}{\cellcolor{LigthGray} --} \\
                        \cline{2-4} \cline{6-9}
                        \multirow{-2}{*}{\textbf{Cryptomining}} & \cellcolor[HTML]{BBFFBB}Our approach  & \cellcolor[HTML]{BBFFBB}\textbf{95.29} & \cellcolor[HTML]{BBFFBB}\textbf{93.11}  & &  \multicolumn{4}{c|}{\cellcolor{LigthGray} --}  \\
                        \cline{1-4} \cline{6-9}
                         & Kumaar et al.~\cite{kumaar2021hybrid} & 99.19  & 99.20   &  & \multicolumn{4}{c|}{\cellcolor{LigthGray} --} \\
                        \cline{2-4} \cline{6-9}
                        \multirow{-2}{*}{\textbf{DNS Malware}} & Our approach & 100.0 & 100.0   & & \multicolumn{4}{c|}{\cellcolor{LigthGray} --} \\
                        \cline{1-4} \cline{6-9}
                          & Zebin et al.~\cite{9796558} & 99.98 & 99.91  &  & \multicolumn{4}{c|}{\cellcolor{LigthGray} --} \\
                        \cline{2-4} \cline{6-9}
                        \multirow{-2}{*}{\textbf{DoH - CIC}} & Our approach  & 99.90 & 99.84   & & \multicolumn{4}{c|}{\cellcolor{LigthGray} --} \\
                        \cline{1-4} \cline{6-9}
                         & Jeřábek et al.~\cite{doh_kamil} &  97.5 &  98.7   & & \multicolumn{4}{c|}{\cellcolor{LigthGray} --} \\ 
                        \cline{2-4} \cline{6-9}
                         \textbf{DoH - Real-world} & Our approach  & 97.79 & 98.80  &  & \multicolumn{4}{c|}{\cellcolor{LigthGray} --} \\
                        \cline{1-4} \cline{6-9}
                        & Shagiq et al.~\cite{SHAFIQ2020433} & 99.99 & 99.99   & & \multicolumn{4}{c|}{\cellcolor{LigthGray} --} \\
                        \cline{2-4} \cline{6-9}
                        \multirow{-2}{*}{\textbf{DoS}} & Our approach &  100.0 & 100.0  &  & \multicolumn{4}{c|}{\cellcolor{LigthGray} --} \\
                        \cline{1-4} \cline{6-9}
                         & \cellcolor[HTML]{BBFFBB}Luxemburk et al.~\cite{9375998} &  \cellcolor[HTML]{BBFFBB}99.93 & \cellcolor[HTML]{BBFFBB}96.26    & & \multicolumn{4}{c|}{\cellcolor{LigthGray} --} \\ 
                        \cline{2-4} \cline{6-9}
                        \multirow{-2}{*}{\textbf{HTTPS Brute-force}} & \cellcolor[HTML]{BBFFBB}Our approach & \cellcolor[HTML]{BBFFBB}\textbf{99.99} & \cellcolor[HTML]{BBFFBB}\textbf{99.83}   &  & \multicolumn{4}{c|}{\cellcolor{LigthGray} --} \\
                        \cline{1-4} \cline{6-9}
                         & \cellcolor[HTML]{BBFFBB}Agrafiotis~\cite{Agrafiotis_Makri_Flionis_Lalas_Votis_Tzovaras_2022} & \cellcolor[HTML]{BBFFBB}98.5 & \cellcolor[HTML]{BBFFBB}95.4 & \cellcolor{white} &  \cellcolor[HTML]{BBFFBB}Kunang et al.~\cite{kunang2021attack} & \cellcolor[HTML]{BBFFBB}Weighted &  \cellcolor[HTML]{BBFFBB}95.79 &  \cellcolor[HTML]{BBFFBB}95.11 \\
                        \cline{2-4} \cline{6-9}
                         & \cellcolor[HTML]{BBFFBB}& \cellcolor[HTML]{BBFFBB}& \cellcolor[HTML]{BBFFBB}& \cellcolor{white} &  \cellcolor[HTML]{BBFFBB} &  \cellcolor[HTML]{BBFFBB}Macro & \cellcolor[HTML]{BBFFBB}\textbf{99.93} & \cellcolor[HTML]{BBFFBB}\textbf{83.23} \\
                         \multirow{-3}{*}{\textbf{IDS - CIC}} & \cellcolor[HTML]{BBFFBB}\multirow{-2}{*}{Our approach} & \cellcolor[HTML]{BBFFBB}\multirow{-2}{*}{\textbf{99.89}} & \cellcolor[HTML]{BBFFBB}\multirow{-2}{*}{\textbf{99.75}} & \cellcolor{white} &  \cellcolor[HTML]{BBFFBB}\multirow{-2}{*}{Our approach}  & \cellcolor[HTML]{BBFFBB}Weighted & \cellcolor[HTML]{BBFFBB}\textbf{99.93} & \cellcolor[HTML]{BBFFBB}\textbf{99.92} \\
                        \cline{1-4} \cline{6-9}
                         & \cellcolor[HTML]{BBFFBB}Ding et al.~\cite{ding2022imbalanced}  & \cellcolor[HTML]{BBFFBB}92.39 &  \cellcolor[HTML]{BBFFBB}94.39 & \cellcolor{white} &  Ding et al. ~\cite{ding2022imbalanced}   & Macro & \cellcolor[HTML]{BBFFBB}90.39 & \cellcolor[HTML]{FAE5E4}\textbf{79.64} \\
                        \cline{2-4} \cline{6-9}
                         &\cellcolor[HTML]{BBFFBB} & \cellcolor[HTML]{BBFFBB} &  \cellcolor[HTML]{BBFFBB} & \cellcolor{white} &  & Macro & \cellcolor[HTML]{BBFFBB}\textbf{95.60} & \cellcolor[HTML]{FAE5E4}40.22 \\
                         \multirow{-3}{*}{\textbf{IDS - UNSW}} & \cellcolor[HTML]{BBFFBB}\multirow{-2}{*}{Our approach}  & \cellcolor[HTML]{BBFFBB}\multirow{-2}{*}{\textbf{98.49}} & \cellcolor[HTML]{BBFFBB}\multirow{-2}{*}{\textbf{98.50}} & \cellcolor{white} & \multirow{-2}{*}{Our approach}  & Weighted & \cellcolor[HTML]{BBFFBB}\textbf{95.60} & \cellcolor[HTML]{FAE5E4}95.08 \\
                        \cline{1-4} \cline{6-9}
                         & Khacha et al.~\cite{khacha2022hybrid}   & 99.99 &  99.99  & & \cellcolor[HTML]{BBFFBB}Khacha et al.~\cite{khacha2022hybrid} & \cellcolor[HTML]{BBFFBB}Weighted  & \cellcolor[HTML]{BBFFBB}98.69 & \cellcolor[HTML]{BBFFBB}-- \\
                        \cline{2-4} \cline{6-9}
                         & & &  & & \cellcolor[HTML]{BBFFBB}  & \cellcolor[HTML]{BBFFBB}Macro & \cellcolor[HTML]{BBFFBB}\textbf{99.97} & \cellcolor[HTML]{BBFFBB}\textbf{89.75} \\
                        \multirow{-3}{*}{\textbf{IoT Malware - Edge-IIoTset}} & \multirow{-2}{*}{Our approach} & \multirow{-2}{*}{99.99} & \multirow{-2}{*}{99.97} &  & \multirow{-2}{*}{\cellcolor[HTML]{BBFFBB}Our approach}  & \cellcolor[HTML]{BBFFBB}Weighted & \cellcolor[HTML]{BBFFBB}\textbf{99.97} &  \cellcolor[HTML]{BBFFBB}\textbf{99.97} \\
                        \cline{1-4} \cline{6-9}
                         & \cellcolor[HTML]{BBFFBB}Sahu et al.~\cite{sahu2021internet} & \cellcolor[HTML]{BBFFBB}$\thicksim 96$ &  \cellcolor[HTML]{BBFFBB}$\thicksim 96$   &  & \multicolumn{4}{c|}{\cellcolor{LigthGray} --} \\
                        \cline{2-4} \cline{6-9}
                         \multirow{-2}{*}{\textbf{IoT Malware - IoT-23}} & \cellcolor[HTML]{BBFFBB}Our approach & \cellcolor[HTML]{BBFFBB}\textbf{99.86} & \cellcolor[HTML]{BBFFBB}\textbf{99.91}   &  & \multicolumn{4}{c|}{\cellcolor{LigthGray} --} \\
                        \cline{1-4} \cline{6-9}
                         & \cellcolor[HTML]{BBFFBB}Dai et al.~\cite{dai2023glads} & \cellcolor[HTML]{BBFFBB}99.29 & \cellcolor[HTML]{BBFFBB}99.03  &  & \cellcolor[HTML]{FAE5E4}Tareq et al. ~\cite{tareq2022analysis} & \cellcolor[HTML]{FAE5E4}Weighted & \cellcolor[HTML]{FAE5E4}98.5 & \cellcolor[HTML]{FAE5E4}\textbf{98.57} \\ 
                        \cline{2-4} \cline{6-9}
                         & \cellcolor[HTML]{BBFFBB} & \cellcolor[HTML]{BBFFBB} & \cellcolor[HTML]{BBFFBB} &  & \cellcolor[HTML]{FAE5E4} & \cellcolor[HTML]{FAE5E4}Macro & \cellcolor[HTML]{FAE5E4}97.53 & \cellcolor[HTML]{FAE5E4}81.02 \\
                        \multirow{-3}{*}{\textbf{IoT Malware - TON\_IoT }} & \cellcolor[HTML]{BBFFBB}\multirow{-2}{*}{Our approach} &  \cellcolor[HTML]{BBFFBB}\multirow{-2}{*}{\textbf{99.96}} & \cellcolor[HTML]{BBFFBB}\multirow{-2}{*}{\textbf{99.98}} &  & \multirow{-2}{*}{\cellcolor[HTML]{FAE5E4}Our approach}  & \cellcolor[HTML]{FAE5E4}Weighted & \cellcolor[HTML]{FAE5E4}97.53 & \cellcolor[HTML]{FAE5E4}97.51 \\
                        \cline{1-4} \cline{6-9}
                         & \cellcolor[HTML]{FAE5E4}Sarkar et al.~\cite{sarkar2020detection} & \cellcolor[HTML]{FAE5E4}99.89 & \cellcolor[HTML]{FAE5E4}\textbf{99.88}  & \cellcolor{white} & Yang et al. ~\cite{yang2023network} & Weighted & 96.04 &  95.97 \\
                        \cline{2-4} \cline{6-9}
                        & \cellcolor[HTML]{FAE5E4} & \cellcolor[HTML]{FAE5E4} & \cellcolor[HTML]{FAE5E4} & \cellcolor{white} & & Macro & 95.48 & 79.87 \\
                        \multirow{-3}{*}{\textbf{TOR}} & \cellcolor[HTML]{FAE5E4}\multirow{-2}{*}{Our approach} & \multirow{-2}{*}{\cellcolor[HTML]{FAE5E4}99.84} & \cellcolor[HTML]{FAE5E4}\multirow{-2}{*}{96.33}  & \cellcolor{white} &  \multirow{-2}{*}{Our approach}  & Weighted & 95.48 & 95.20 \\
                        \cline{1-4} \cline{6-9}
                        \rowcolor{white} & \cellcolor[HTML]{BBFFBB}Aceto  et  al.~\cite{ACETO2021102985}  & \cellcolor[HTML]{BBFFBB}93.75  & \cellcolor[HTML]{BBFFBB}91.95  &  & \cellcolor[HTML]{BBFFBB}Dener et al.  ~\cite{dener2023rfse} &  \cellcolor[HTML]{BBFFBB}Macro & \cellcolor[HTML]{BBFFBB}89.29 & \cellcolor[HTML]{BBFFBB}87.83  \\
                        \cline{2-4} \cline{6-9}
                         & \cellcolor[HTML]{BBFFBB} & \cellcolor[HTML]{BBFFBB} & \cellcolor[HTML]{BBFFBB} & & \cellcolor[HTML]{BBFFBB}  &  \cellcolor[HTML]{BBFFBB}Macro &  \cellcolor[HTML]{BBFFBB}\textbf{94.80} &  \cellcolor[HTML]{BBFFBB}\textbf{91.21} \\
                        \multirow{-3}{*}{\textbf{VPN - ISCX}}  & \multirow{-2}{*}{\cellcolor[HTML]{BBFFBB}Our approach} &  \multirow{-2}{*}{\cellcolor[HTML]{BBFFBB}\textbf{94.35}} & \multirow{-2}{*}{\cellcolor[HTML]{BBFFBB}\textbf{95.48}} &  & \multirow{-2}{*}{\cellcolor[HTML]{BBFFBB}Our approach} & \cellcolor[HTML]{BBFFBB}Weighted & \cellcolor[HTML]{BBFFBB}\textbf{94.80} & \cellcolor[HTML]{BBFFBB}\textbf{94.77} \\
                        \cline{1-4} \cline{6-9}
                         & \cellcolor[HTML]{BBFFBB}Jorgense et al.~\cite{vnet_dataset}   &  \cellcolor[HTML]{BBFFBB}-- & \cellcolor[HTML]{BBFFBB}98.00  & \cellcolor{white} &  \cellcolor[HTML]{BBFFBB}Jorgensen et al. ~\cite{vnet_dataset} & \cellcolor[HTML]{BBFFBB}Macro & \cellcolor[HTML]{BBFFBB}96 & \cellcolor[HTML]{BBFFBB}-- \\
                        \cline{2-4} \cline{6-9}
                         & \cellcolor[HTML]{BBFFBB}& \cellcolor[HTML]{BBFFBB}& \cellcolor[HTML]{BBFFBB} & \cellcolor{white} &  \cellcolor[HTML]{BBFFBB}& \cellcolor[HTML]{BBFFBB}Macro & \cellcolor[HTML]{BBFFBB}\textbf{98.60} &   \cellcolor[HTML]{BBFFBB}\textbf{98.88} \\
                         \multirow{-3}{*}{\textbf{VPN - VNAT}}  & \cellcolor[HTML]{BBFFBB}\multirow{-2}{*}{Our approach} & \cellcolor[HTML]{BBFFBB}\multirow{-2}{*}{\textbf{99.98}} & \cellcolor[HTML]{BBFFBB}\multirow{-2}{*}{\textbf{99.72}} & \cellcolor{white} & \cellcolor[HTML]{BBFFBB}\multirow{-2}{*}{Our approach}  & \cellcolor[HTML]{BBFFBB}Weighted & \cellcolor[HTML]{BBFFBB}\textbf{98.60} & \cellcolor[HTML]{BBFFBB}\textbf{98.60} \\
                        \cline{1-4} \cline{6-9}
                    \end{tabular}
                    \label{evaluation}
                \end{center}
            \end{table*}

\section{Feature Evaluation} \label{classification_section}
    We evaluated the features by creating a novel classifier for each concerned network classification task. The classifier creation pipeline is the set of steps that creates the best final model. At first, the published datasets were split among Train, Validation, and Test sets in a ratio of 60:20:20 while keeping the labeling ratio like in the original datasets. Furthermore, some additional value sanitation is recommended, e.g., for a very short time series, it is required to handle ``NaN'' values:  we replace NaN for the distribution features with 0.5, for the frequency features with -1, and for the rest of the features with 0. The source codes of our whole classification pipeline, including the pre-processing, are available at Github\footnote{\url{https://github.com/koumajos/ClassificationBasedOnSFTS}}.
                
    In the validation phase, we first select the optimal ML algorithm. We test 14 well-known ML algorithms such as Random Forest, K-NN or SVM; nevertheless, the XGBoost algorithm achieved the best performance among all of the evaluated classifications tasks. After the algorithm selection, we searched for the best model hyperparameters to get optimal performance on each dataset without overfitting. We use the \texttt{hyperopt library}~\cite{bergstra2013making} to tune the following hyperparameters: \texttt{n\_estimators}, \texttt{max\_depth}, \texttt{gamma}, \texttt{reg\_alpha}, \texttt{min\_child\_weight}, and \texttt{colsample\_bytree}.  

    The hyperparameter search was performed using the training and validation datasets. The best values of the hyperparameters were selected based on the \textit{F1-score} measure on the validation dataset. The final performance of the classifier on each dataset was obtained from the model trained using the trained part and evaluated on the test part. The test part was not used during any stage of the classifier design, ensuring the fairness of model evaluation on data that was not seen before.

    \subsection{Results}
        The results of binary and multiclass classification are presented in Table~\ref{evaluation}. On most of the binary classification problems, the novel feature set achieved similar or better performance than the best-performing previous work. Moreover, our approach outperformed eight related works significantly (by more than 1\%). However, on the TOR detection problem, we obtained a worse F1-score than the best classifier. 

        We investigated the differences between the TOR classifier published by Sarkar et al.~\cite{sarkar2020detection} and found out that he uses a specially tailored feature vector that also include transport ports. Even though transport ports are often used in network traffic classification, we intentionally opted to avoid them to maintain the universality of features. The classifier often tends to overfit the transport port features, which, in some cases, is not a desired behavior. 

        When we analyze the performance of the multiclass classification, we also outperformed most of the best-performing classifiers. Specifically, in five out of eight cases, we achieved more than a 1\% classification performance increase. However, in two cases, we observed a slight decrease---TON\_IoT and IDS-UNSW cases. 

        The best-performing classifier of TON\_IoT published by Tareq et al.~\cite{tareq2022analysis} is based on a 2D convolutional network (CNN) with very long packet-length data (SPLT with all packets from connection) organized in the image. The SPLT data give the classifier advantage in the opportunity of high-quality feature extraction that allows accurate classification. Nevertheless, the long packet sequences (SPLT) cannot be exported in real-world deployment scenarios due to the technical limitations of the flow exporters (see Section~\ref{related_works_section}).

        Besides, Ding et al.~\cite{ding2022imbalanced} achieved better results with IDS classification using IDS - UNSW dataset. According to our analysis, the better results are caused by a high-class imbalance of the dataset. Ding et al. thus proposed techniques for dealing with the imbalance ratio between classes. In our case, we did not deploy any imbalanced learn techniques in our classifier design pipeline to maintain comparability with the previous works---most of the concerned related works do not deploy any imbalanced learn techniques. 

        As can be seen in Table~\ref{evaluation}, the proposed feature vector proved to be universal and performed well on all concerned tasks. The slight reduction in accuracy in some cases was expected since universal features cannot compete with specially tailored ones; however, we still consider the performance reduction, especially in the case of TOR or TON\_IoT as a good tradeoff for universality and the possibility of deploying all the network classifiers behind single flow monitoring device. 

\subsection{Feature importance}         
\begin{figure}[ht]
            \centerline{\includegraphics[width=8cm]{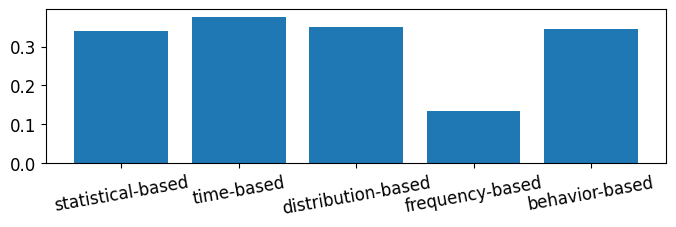}}
            \caption{Average feature importance of each feature category}
            \label{fig:feature_set_2}
        \end{figure}

\begin{figure*}
            \centerline{\includegraphics[width=18cm]{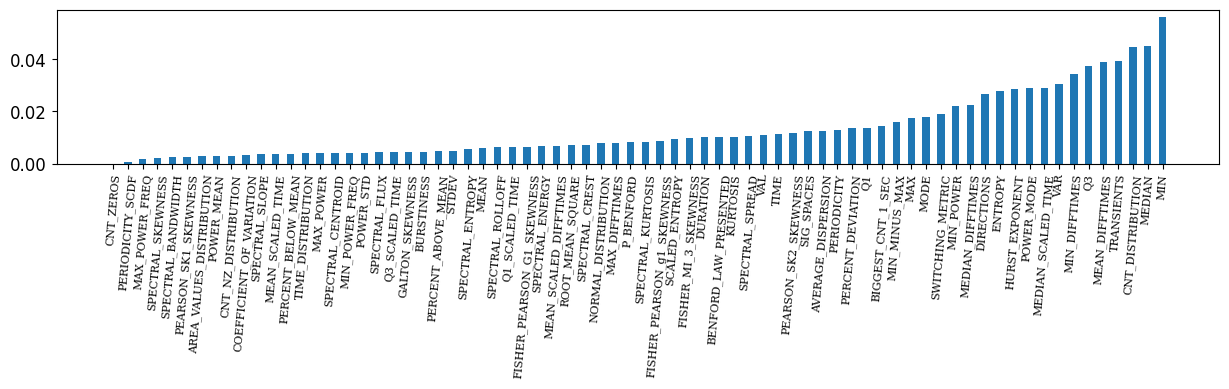}}
            \caption{Average feature importance of presented feature set}
            \label{fig:feature_set}
    \end{figure*}

       We present the average feature importance in Fig. \ref{fig:feature_set}. On average, the best features are basic statistical features and, surprisingly, the Transient feature which is important mainly in Botnet and IoT malware detection. Furthermore, we compute the average feature importance per each category of features. Fig. \ref{fig:feature_set_2} shows that most categories have equal average feature importance. Only the frequency-based category has a minor significance in the average. However, one of the frequency-based features, \textit{Min power}, is in the top ten best features. 

\section{Reduced Featureset} \label{feature_reduction_section}
    Network telemetry plays an important role in classifying network traffic. It is not enough to have a classification method that has excellent classification results. There is also the need to consider the network load of telemetry required for classification. Creating a balanced method in the size of network telemetry and classification results is necessary for practical reasons. 
    
    For this purpose, we experimented with feature set reduction to find an ideally small number of features that perform well, by comparing the loss or gain of classification results. Furthermore, we must reduce the feature set to a subset suitable to all presented problems, even for multiclass problems.

    We evaluated the feature importance on all 23 models, that were presented in this paper. Most of the binary problems use mainly statistical-based features. On the contrary, the frequency-based and distribution-based features are the least important, with approximately the same results. However, for multiclass classification, the frequency and distribution features play an important role.

    From the feature importance evaluation, we obtain 10 unique feature sets; each one containing 10 features. These feature sets were used as input for the classification pipeline. Thus, the results were computed using each binary/multiclass problem and feature set, i.e.,  23 * 10 = 230 best models were created and evaluated\footnote{The total number of models evaluated in our validation phase for picking the best model was even higher, in order of magnitude 2,500.}. Based on the results, we compare the 10 feature sets using the statistical average and standard deviation. The best feature set achieved similar results as the original feature set with 69 features. 

    The selected best feature set contains the following features sorted by feature importance:  \textit{Spectral kurtosis}, \textit{Periodicity}, \textit{First quartile}, \textit{Benford's law}, \textit{Spectral energy}, \textit{Median of time differences}, \textit{Min}, \textit{Third quartile}, \textit{Min minus max}, and \textit{Directions}.

    This feature set was combined with fields from classic IP flow that are not network or device dependent, i.e., flow duration and the number of packets, bytes, packets in the reverse direction, and bytes in the reverse direction. The resulting classification performance reduction  is shown in Table~\ref{tab:feature_reduction_average_best_10}. The main loss of classification results is caused by the multiclass classification of Tor traffic, where the feature set performs significantly worse. Nevertheless, the average accuracy reduction across all use cases is below 0.11\%. 

        \begin{table}[ht]
              \caption{Average loss of classification Accuracy and F1-score (in \%) of selected set of features.}
            \begin{center}
                \begin{tabular}{|R|R|R|R|R|}
                    \hline
                    \rowcolor{Gray}  & \textbf{Accuracy} &
                    \textbf{F1-score} \\
                    \hline
                    \rowcolor{white} \textbf{Average} & -0.03 (\(\pm\) 0.70)  & -0.08 (\(\pm\) 2.01) \\
                    \hline
                    \textbf{Average binary} & -0.11 (\(\pm\) 0.29) & -0.09 (\(\pm\) 0.24) \\
                    \hline
                    \rowcolor{white} \textbf{Average macro} & 0.02 (\(\pm\) 1.32)  & 0.06 (\(\pm\) 3.60) \\
                    \hline
                    \textbf{Average weighed} &  0.02 (\(\pm\) 1.32) &  -0.19 (\(\pm\) 0.99) \\
                    \hline
                \end{tabular}
                \label{tab:feature_reduction_average_best_10}
            \end{center}
        \end{table}

\section{Conclusion}\label{conclusion_section}
    In this paper, we propose a novel feature set built from Time Series Analysis of Single Flow Time Series that can be used for classification methods. The proposed feature set is highly universal and achieves great results for both binary and multiclass classification. The feature set covers a wide range of behavior types in the following groups: 1)~statistical deviation of payload lengths, 2)~statistical deviation of packets times, 3)~distribution of packets, 4)~behavior of frequency domain, and 5)~specific behaviors of data points. All groups contain significant features for classification.

    The proposed method and feature set was evaluated on 23 network classification tasks using 15 publicly available and well-known network traffic datasets which are often used in recent research. All the collected datasets were processed to compute the proposed time series features that were published for any further research by the scientific community. Additionally, we also identified a reduced feature set of the 10 most significant features to allow deployment in monitoring infrastructures with strict bandwidth constraints. The reduced feature set of only 10 features naturally comes with the cost of reduced classification performance. However, the average accuracy drop across the whole task was below 0.11\%

    Overall, we trained and evaluated over 2500 models across multiple binary and multiclass classification tasks and showed the universality of the proposed features. Furthermore, we prepared the highly efficient C++ implementation of the proposed feature vector extraction methods inside flow exporter ipfixprobe\footnote{\url{https://github.com/koumajos/ipfixprobe_tsa_sfts}}. The developed library allows a fast integration into other flow-monitoring software.
    


\bibliographystyle{unsrt}
\bibliography{bibliography}

\begin{thebibliography}{10}

\bibitem{ietf-tls-esni-16}
Eric Rescorla et~al.
\newblock {TLS Encrypted Client Hello}.
\newblock Internet-Draft draft-ietf-tls-esni-16, Internet Engineering Task
  Force, 2023.

\bibitem{IP_flows}
Ganesh Sadasivan et~al.
\newblock {A}rchitecture for {IP} {F}low {I}nformation {E}xport.
\newblock {\em {RFC}}, 5470:1--31, 2009.

\bibitem{9892293}
Miguel Leon et~al.
\newblock Comparative evaluation of machine learning algorithms for network
  intrusion detection and attack classification.
\newblock In {\em IJCNN}, pages 01--08, 2022.

\bibitem{stergiopoulos2018automatic}
George Stergiopoulos et~al.
\newblock {A}utomatic {D}etection of {V}arious {M}alicious {T}raffic {U}sing
  {S}ide {C}hannel {F}eatures on {TCP} {P}ackets.
\newblock In {\em {ESORICS} 2018}, volume 11098, pages 346--362. Springer,
  2018.

\bibitem{8844609}
Gonzalo Marín et~al.
\newblock Deep in the dark - deep learning-based malware traffic detection
  without expert knowledge.
\newblock In {\em SPW 2019}, 2019.

\bibitem{wang2017malware}
Wei Wang et~al.
\newblock {M}alware {T}raffic {C}lassification {U}sing {C}onvolutional {N}eural
  {N}etwork {F}or {R}epresentation {L}earning.
\newblock In {\em {ICOIN} 2017}, pages 712--717. {IEEE}, 2017.

\bibitem{plny2023decrypto}
Richard Pln{\'{y}} et~al.
\newblock {D}e{C}rypto: {F}inding {C}ryptocurrency {M}iners on {ISP}
  {N}etworks.
\newblock In {\em NordSec 2022}. Springer, 2022.

\bibitem{montazerishatoori2020detection}
Mohammadreza MontazeriShatoori et~al.
\newblock Detection of doh tunnels using time-series classification of
  encrypted traffic.
\newblock In {\em DASC/PiCom/CBDCom/CyberSciTech 2020}, pages 63--70. IEEE,
  2020.

\bibitem{doh_kamil}
Kamil Jerabek, Karel Hynek, Ondrej Rysavy, and Ivana Burgetova.
\newblock Dns over https detection using standard flow telemetry.
\newblock {\em IEEE Access}, 11:50000--50012, 2023.

\bibitem{9375998}
Jan Luxemburk et~al.
\newblock Detection of https brute-force attacks with packet-level feature set.
\newblock In {\em CCWC 2021}, pages 0114--0122, 2021.

\bibitem{luxemburk2022fine}
Jan Luxemburk et~al.
\newblock {F}ine-grained {TLS} {S}ervices {C}lassification {W}ith {R}eject
  {O}ption.
\newblock {\em Comput. Networks}, 220:109467, 2023.

\bibitem{sblt_our_paper}
Zdena Tropkov{\'{a}} et~al.
\newblock {N}ovel {HTTPS} {C}lassifier {D}riven by {P}acket {B}ursts, {F}lows,
  and {M}achine {L}earning.
\newblock In {\em {CNSM} 2021}. {IEEE}, 2021.

\bibitem{lashkari2017characterization}
Arash~Habibi Lashkari et~al.
\newblock {C}haracterization of {T}or {T}raffic using {T}ime based {F}eatures.
\newblock In {\em {ICISSP} 2017}, pages 253--262. SciTePress, 2017.

\bibitem{http_extend_ip_flow}
Martin Hus{\'{a}}k et~al.
\newblock {S}ecurity {M}onitoring of {HTTP} {T}raffic {U}sing {E}xtended
  {F}lows.
\newblock In {\em {ARES} 2015}. {IEEE} Computer Society.

\bibitem{6356064}
Min Hur et~al.
\newblock Towards smart phone traffic classification.
\newblock In {\em APNOMS}, pages 1--4, 2012.

\bibitem{bayat2021deep}
Niloofar Bayat et~al.
\newblock Deep learning for network traffic classification, 2021.

\bibitem{josef_koumar_2023_8035724}
Josef Koumar, Karel Hynek, and Tomáš Čejka.
\newblock {Network traffic datasets created by Single Flow Time Series
  Analysis}, 2023.

\bibitem{Jerabek2022}
Kamil Jeřábek et~al.
\newblock {C}ollection of datasets with {D}{N}{S} over {H}{T}{T}{P}{S} traffic.
\newblock {\em Data in Brief}, 42:108310, 2022.

\bibitem{vekshin2020doh}
Dmitrii Vekshin et~al.
\newblock {D}o{H} {I}nsight: {D}etecting {DNS} {O}ver {HTTPS} by {M}achine
  {L}earning.
\newblock In {\em {ARES} 2020}. {ACM}, 2020.

\bibitem{Hofstede2017}
Rick Hofstede, Mattijs Jonker, Anna Sperotto, and Aiko Pras.
\newblock Flow-based web application brute-force attack and compromise
  detection.
\newblock {\em Journal of Network and Systems Management}, August 2017.

\bibitem{9796558}
Tahmina Zebin et~al.
\newblock An explainable ai-based intrusion detection system for dns over https
  (doh) attacks.
\newblock {\em IEEE Transactions on Information Forensics and Security},
  17:2339--2349, 2022.

\bibitem{cic_bell_dns_2021_article}
Samaneh Mahdavifar et~al.
\newblock {C}lassifying {M}alicious {D}omains using {DNS} {T}raffic {A}nalysis.
\newblock In {\em DASC/PiCom/CBDCom/CyberSciTech 2021}, pages 60--67. {IEEE},
  2021.

\bibitem{sharafaldin2018toward}
Iman Sharafaldin et~al.
\newblock Toward generating a new intrusion detection dataset and intrusion
  traffic characterization.
\newblock {\em ICISSp}, 1:108--116, 2018.

\bibitem{Agrafiotis_Makri_Flionis_Lalas_Votis_Tzovaras_2022}
Georgios Agrafiotis et~al.
\newblock Image-based neural network models for malware traffic classification
  using pcap to picture conversion.
\newblock In {\em Proceedings of the 17th International Conference on
  Availability, Reliability and Security}, pages 1--7, 2022.

\bibitem{ding2022imbalanced}
Hongwei Ding et~al.
\newblock Imbalanced data classification: A knn and generative adversarial
  networks-based hybrid approach for intrusion detection.
\newblock {\em Future Generation Computer Systems}, 131:240--254, 2022.

\bibitem{DoH_cic_dataset}
Mohammadreza MontazeriShatoori et~al.
\newblock {D}etection of {D}o{H} {T}unnels using {T}ime-series {C}lassification
  of {E}ncrypted {T}raffic.
\newblock In {\em DASC/PiCom/CBDCom/CyberSciTech 2020}, pages 63--70. {IEEE},
  2020.

\bibitem{koumar2022network}
Josef Koumar et~al.
\newblock {N}etwork {T}raffic {C}lassification {B}ased on {P}eriodic {B}ehavior
  {D}etection.
\newblock In {\em {CNSM} 2022}, pages 359--363. {IEEE}, 2022.

\bibitem{KoumarUSTS}
Josef Koumar and Tomáš Čejka.
\newblock {U}nevenly {S}paced {T}ime {S}eries from {N}etwork {T}raffic.
\newblock {\em TMA}, 2023.
\newblock preprint on webpage at
  \url{https://www.researchgate.net/publication/371530461_Unevenly_Spaced_Time_Series_from_Network_Traffic}.

\bibitem{SZABO200497}
Thomas~L. Szabo.
\newblock 5 - {T}ransducers.
\newblock In Thomas~L. Szabo, editor, {\em Diagnostic Ultrasound Imaging},
  Biomedical Engineering, pages 97--135. Academic Press, Burlington, 2004.

\bibitem{scheirer1997construction}
Eric~D. Scheirer et~al.
\newblock {C}onstruction and {E}valuation of a {R}obust {M}ultifeature
  {S}peech/{M}usic {D}iscriminator.
\newblock In {\em {ICASSP} 1997}, pages 1331--1334. {IEEE} Computer Society,
  1997.

\bibitem{lerch2012introduction}
Alexander Lerch.
\newblock {\em {A}n {I}ntroduction to {A}udio {C}ontent {A}nalysis:
  {A}pplications in {S}ignal {P}rocessing and {M}usic {I}nformatics}.
\newblock Wiley-IEEE Press, 2012.

\bibitem{hurst_exponent}
A.~A. Anis et~al.
\newblock {T}he {E}xpected {V}alue of the {A}djusted {R}escaled {H}urst {R}ange
  of {I}ndependent {N}ormal {S}ummands.
\newblock {\em Biometrika}, 63(1), 1976.

\bibitem{miller2015benford}
Steven~J Miller.
\newblock {\em {B}enford's {L}aw}.
\newblock Princeton University Press, 2015.

\bibitem{fda_tda_comparsion_1}
A.~Suarez et~al.
\newblock {A}nalytical {C}omparison {B}etween {T}ime- {A}nd {F}requency-domain
  {T}echniques for {P}hase-noise {A}nalysis.
\newblock {\em IEEE Transactions on Microwave Theory and Techniques},
  50(10):2353--2361, 2002.

\bibitem{fda_tda_comparsion_2}
S.~J Worley et~al.
\newblock {C}omparison of {T}ime {D}omain and {F}requency {D}omain {V}ariables
  {F}rom the {S}ignal-averaged {E}lectrocardiogram: {A} {M}ultivariable
  {A}nalysis.
\newblock {\em Journal of the American College of Cardiology}, 1988.

\bibitem{fda_tda_comparsion_3}
Ralph Haberl et~al.
\newblock {C}omparison of {F}requency and {T}ime {D}omain {A}nalysis of the
  {S}ignal-averaged {E}lectrocardiogram in {P}atients {W}ith {V}entricular
  {T}achycardia and {C}oronary {A}rtery {D}isease: {M}ethodologic {V}alidation
  and {C}linical {R}elevance.
\newblock {\em Journal of the American College of Cardiology}, 12(1):150--158,
  1988.

\bibitem{lomb}
Nicholas~R Lomb.
\newblock Least-squares frequency analysis of unequally spaced data.
\newblock {\em Astrophysics and space science}, 39:447--462, 1976.

\bibitem{scargle}
Jeffrey Scargle.
\newblock {S}tudies in astronomical time series analysis. {I}{I} -
  {S}tatistical aspects of spectral analysis of unevenly spaced data.
\newblock {\em The Astrophysical Journal}, 263, December 1982.

\bibitem{LS_periodogram}
Jacob~T. VanderPlas.
\newblock {U}nderstanding the lomb{\textendash}scargle {P}eriodogram.
\newblock {\em The Astrophysical Journal Supplement Series}, 236(1):16, may
  2018.

\bibitem{boashash2015time}
Boualem Boashash.
\newblock {\em {T}ime-Frequency {S}ignal {A}nalysis and {P}rocessing: {A}
  {C}omprehensive {R}eference}.
\newblock Academic press, 2015.

\bibitem{spectral_kurtosis_compute}
Valeriu Vrabie et~al.
\newblock {S}pectral {K}urtosis: {F}rom {D}efinition to {A}pplication.
\newblock In {\em NSIP 2003}, 2003.

\bibitem{about_scdf}
FAM Frescura et~al.
\newblock {S}ignificance {T}ests for {P}eriodogram {P}eaks.
\newblock {\em arXiv preprint arXiv:0706.2225}, 2007.

\bibitem{spectral_rolloff_zero_cros}
Md~Gulzar Hussain et~al.
\newblock Classification of bangla alphabets phoneme based on audio features
  using mlpc \& svm.
\newblock In {\em ACMI 2021}. IEEE, 2021.

\bibitem{peeters2004large}
Geoffroy Peeters.
\newblock {A} {L}arge {S}et of {A}udio {F}eatures for {S}ound {D}escription
  ({S}imilarity and {C}lassification) in the {C}{U}{I}{D}{A}{D}{O} {P}roject.
\newblock {\em CUIDADO Ist Project Report}, 54(0):1--25, 2004.

\bibitem{GARCIA2014100}
S.~García et~al.
\newblock {A}n {E}mpirical {C}omparison of {B}otnet {D}etection {M}ethods.
\newblock {\em Computers \& Security}, 45:100--123, 2014.

\bibitem{richard_plny_2022_7189293}
Richard Plný et~al.
\newblock {\em Datasets of Cryptomining Communication}.
\newblock Zenodo, October 2022.

\bibitem{kumaar2021hybrid}
M~Kumaar et~al.
\newblock {A} {H}ybrid {F}ramework for {I}ntrusion {D}etection in {H}ealthcare
  {S}ystems {U}sing {D}eep {L}earning.
\newblock {\em Frontiers in Public Health}, 9, 2021.

\bibitem{jan_luxemburk_2020_4275775}
Jan Luxemburk et~al.
\newblock {HTTPS Brute-force dataset with extended network flows}, November
  2020.

\bibitem{SHAFIQ2020433}
Muhammad Shafiq et~al.
\newblock {S}election of {E}ffective {M}achine {L}earning {A}lgorithm and
  {B}ot-{I}o{T} {A}ttacks {T}raffic {I}dentification for {I}nternet of {T}hings
  in {S}mart {C}ity.
\newblock {\em Future Gener. Comput. Syst.}, 107:433--442, 2020.

\bibitem{dos_iot_Dataset}
Nickolaos Koroniotis et~al.
\newblock {T}owards the development of realistic botnet dataset in the
  {I}nternet of {T}hings for network forensic analytics: {B}ot-{I}o{T} dataset.
\newblock {\em Future Gener. Comput. Syst.}, 100:779--796, 2019.

\bibitem{SHAFIQ2020101863}
Muhammad Shafiq et~al.
\newblock {I}o{T} {M}alicious {T}raffic {I}dentification {U}sing
  {W}rapper-based {F}eature {S}election {M}echanisms.
\newblock {\em Comput. Secur.}, 2020.

\bibitem{khacha2022hybrid}
Amina Khacha et~al.
\newblock Hybrid deep learning-based intrusion detection system for industrial
  internet of things.
\newblock In {\em ISIA}. IEEE, 2022.

\bibitem{mbc1-1h68-22}
Mohamed~Amine Ferrag et~al.
\newblock Edge-iiotset: A new comprehensive realistic cyber security dataset of
  iot and iiot applications: Centralized and federated learning, 2022.

\bibitem{9751703}
Mohamed~Amine Ferrag et~al.
\newblock Edge-iiotset: A new comprehensive realistic cyber security dataset of
  iot and iiot applications for centralized and federated learning.
\newblock {\em IEEE Access}, 10:40281--40306, 2022.

\bibitem{tareq2022analysis}
Imad Tareq et~al.
\newblock Analysis of ton-iot, unw-nb15, and edge-iiot datasets using dl in
  cybersecurity for iot.
\newblock {\em Applied Sciences}, 12(19):9572, 2022.

\bibitem{sahu2021internet}
Amiya~Kumar Sahu et~al.
\newblock Internet of things attack detection using hybrid deep learning model.
\newblock {\em Computer Communications}, 176:146--154, 2021.

\bibitem{sebastian_garcia_2020_4743746}
Sebastian Garcia et~al.
\newblock {IoT-23: A labeled dataset with malicious and benign IoT network
  traffic}, January 2020.

\bibitem{nascita2022machine}
Alfredo Nascita et~al.
\newblock Machine and deep learning approaches for iot attack classification.
\newblock In {\em IEEE INFOCOM}, pages 1--6. IEEE, 2022.

\bibitem{dai2023glads}
Jianbang Dai et~al.
\newblock Glads: A global-local attention data selection model for multimodal
  multitask encrypted traffic classification of iot.
\newblock {\em Computer Networks}, 225:109652, 2023.

\bibitem{moustafa2021new}
Nour Moustafa.
\newblock A new distributed architecture for evaluating ai-based security
  systems at the edge: Network ton\_iot datasets.
\newblock {\em Sustainable Cities and Society}, 72:102994, 2021.

\bibitem{guo2021machine}
Ge~Guo.
\newblock A machine learning framework for intrusion detection system in iot
  networks using an ensemble feature selection method.
\newblock In {\em IEEE IEMCON}, pages 0593--0599. IEEE, 2021.

\bibitem{moustafa2015unsw}
Nour Moustafa et~al.
\newblock Unsw-nb15: a comprehensive data set for network intrusion detection
  systems (unsw-nb15 network data set).
\newblock In {\em MilCIS}, pages 1--6. IEEE, 2015.

\bibitem{mulyanto2020effectiveness}
Mulyanto Mulyanto et~al.
\newblock Effectiveness of focal loss for minority classification in network
  intrusion detection systems.
\newblock {\em Symmetry}, 13(1):4, 2020.

\bibitem{sarkar2020detection}
Debmalya Sarkar et~al.
\newblock {D}etection of {T}or {T}raffic using {D}eep {L}earning.
\newblock In {\em {AICCSA} 2020}, pages 1--8. {IEEE}, 2020.

\bibitem{yang2023network}
Yang Yang et~al.
\newblock A network traffic classification method based on dual-mode feature
  extraction and hybrid neural networks.
\newblock {\em IEEE Transactions on Network and Service Management}, 2023.

\bibitem{cuzzocrea2017tor}
Alfredo Cuzzocrea et~al.
\newblock {T}or {T}raffic {A}nalysis and {D}etection via {M}achine {L}earning
  {T}echniques.
\newblock In {\em {IEEE} BigData 2017}, 2017.

\bibitem{ACETO2021102985}
Giuseppe Aceto et~al.
\newblock {DISTILLER:} {E}ncrypted {T}raffic {C}lassification via {M}ultimodal
  {M}ultitask {D}eep {L}earning.
\newblock {\em J. Netw. Comput. Appl.}, 2021.

\bibitem{dener2023rfse}
Murat Dener et~al.
\newblock Rfse-gru: Data balanced classification model for mobile encrypted
  traffic in big data environment.
\newblock {\em IEEE Access}, 11:21831--21847, 2023.

\bibitem{icxs_vpn_2016_dataset}
Gerard Draper-Gil et~al.
\newblock {C}haracterization of {E}ncrypted and {V}{P}{N} {T}raffic {U}sing
  {T}ime-related.
\newblock In {\em ICISSP}, pages 407--414, 2016.

\bibitem{shapira2019flowpic}
Tal Shapira et~al.
\newblock Flowpic: Encrypted internet traffic classification is as easy as
  image recognition.
\newblock In {\em IEEE INFOCOM 2019}. IEEE, 2019.

\bibitem{vnet_dataset}
Steven Jorgensen et~al.
\newblock {E}xtensible {M}achine {L}earning for {E}ncrypted {N}etwork {T}raffic
  {A}pplication {L}abeling via {U}ncertainty {Q}uantification.
\newblock {\em CoRR}, abs/2205.05628, 2022.

\bibitem{10029386}
John~T. Holodnak et~al.
\newblock {B}ackdoor {P}oisoning of {E}ncrypted {T}raffic {C}lassifiers.
\newblock In {\em {ICDM} 2022}, pages 577--585. {IEEE}, 2022.

\bibitem{behnke2021feature}
Matthew Behnke et~al.
\newblock Feature engineering and machine learning model comparison for
  malicious activity detection in the dns-over-https protocol.
\newblock {\em IEEE Access}, 9:129902--129916, 2021.

\bibitem{kunang2021attack}
Yesi~Novaria Kunang et~al.
\newblock Attack classification of an intrusion detection system using deep
  learning and hyperparameter optimization.
\newblock {\em Journal of Information Security and Applications}, 58:102804,
  2021.

\bibitem{bergstra2013making}
James Bergstra et~al.
\newblock Making a science of model search: Hyperparameter optimization in
  hundreds of dimensions for vision architectures.
\newblock In {\em International conference on machine learning}, pages
  115--123. PMLR, 2013.

\end{thebibliography}

\end{document}